\documentclass[letterpaper, 10 pt, conference]{ieeeconf}  

\IEEEoverridecommandlockouts                              

\usepackage{booktabs}
\usepackage{multirow}
\usepackage{graphicx}
\usepackage{caption}
\usepackage{balance}
\usepackage[backend=biber,
            hyperref=true,
            url=false,
            isbn=false,
            doi=false,
            backref=false,
            style=ieee,
            citestyle=numeric-comp,
            sorting=none,
            block=none]{biblatex}
\usepackage[font=small,labelfont=bf]{caption}
\usepackage{xspace}
\usepackage{xcolor}
\usepackage[bookmarks=true]{hyperref}
\usepackage{cleveref}
\definecolor{navyblue}{RGB}{0,0,128}
\hypersetup{
  colorlinks = true, 
  urlcolor = navyblue, 
  linkcolor = navyblue, 
  citecolor = navyblue, 
}

\usepackage{xspace}
\newcommand{\model}[0]{\textsc{\textbf{Lidea}}\xspace}

\newcommand{\qheading}[1]{\vspace{2pt}\textbf{#1}\xspace}

\def\github{\raisebox{-1.5pt}{\includegraphics[height=1.0em]{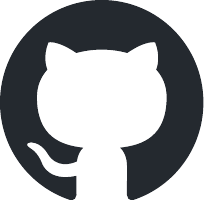}}}

\title{\LARGE \bf
\textsc{\model}: Human-to-Robot Imitation \underline{L}earning via \\ \underline{I}mplicit Feature \underline{D}istillation and \underline{E}xplicit Geometry \underline{A}lignment
}

\author{
\authorblockN{
Yifu Xu$^{12\star}$,
Bokai Lin$^{12\star}$,
Xinyu Zhan$^{12}$,
Hongjie Fang$^{1}$,
Yong-Lu Li$^{12}$,
Cewu Lu$^{123}$,
Lixin Yang$^{123\dagger}$
}
\authorblockA{
$^{1}$Shanghai Jiao Tong University \quad
$^{2}$Shanghai Innovation Institute \quad
$^{3}$Noematrix Ltd.
}
\thanks{$^{\star}$Equal contribution. $^{\dagger}$Corresponding author.}%
}

\begin{document}

\makeatletter
\let\@oldmaketitle\@maketitle
\renewcommand{\@maketitle}{
\@oldmaketitle
\vspace{0.2cm}
\centering
\begin{minipage}{\textwidth}
    \centering
    \includegraphics[width=\linewidth]{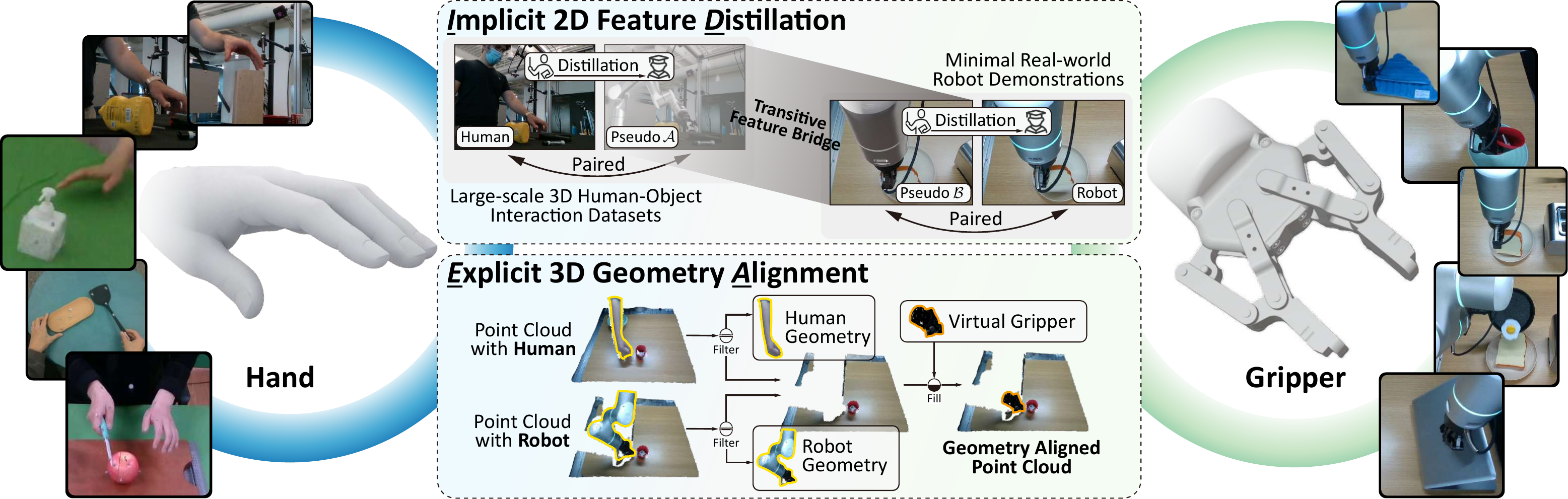}
    \captionof{figure}{\textbf{Overview of \model.} \model bridges the embodiment gap between human hands and robot grippers from two complementary aspects: \textit{(Top)} implicit 2D feature distillation utilizes a transitive feature bridge to align human and robot representations; \textit{(Bottom)} explicit 3D geometry alignment filters embodiment-specific geometries and fills a virtual gripper to construct a geometry-aligned point cloud.}
    \label{fig:teaser}

    {\raggedright
    \small \github~\textbf{Project Page}: \href{https://yifuxu1127.github.io/LIDEA}{yifuxu1127.github.io/LIDEA}\par
    }
\end{minipage}

}
\makeatother

\maketitle
\thispagestyle{plain}
\pagestyle{plain}
\addtocounter{figure}{-1}

\begin{abstract}
Scaling up robot learning is hindered by the scarcity of robotic demonstrations, whereas human videos offer a vast, untapped source of interaction data. However, bridging the embodiment gap between human hands and robot arms remains a critical challenge. Existing cross-embodiment transfer strategies typically rely on visual editing, but they often introduce visual artifacts due to intrinsic discrepancies in visual appearance and 3D geometry. To address these limitations, we introduce \textbf{\model} (Implicit Feature Distillation and Explicit Geometric Alignment), an imitation learning framework in which policy learning benefits from human demonstrations. In the 2D visual domain, \textbf{\model} employs a dual-stage transitive distillation pipeline that aligns human and robot representations in a shared latent space. In the 3D geometric domain, we propose an embodiment-agnostic alignment strategy that explicitly decouples embodiment from interaction geometry, ensuring consistent 3D-aware perception. Extensive experiments empirically validate \textbf{\model} from two perspectives: data efficiency and OOD robustness. Results show that human data substitutes up to 80\% of costly robot demonstrations, and the framework successfully transfers unseen patterns from human videos for out-of-distribution generalization. 
\end{abstract}

\section{Introduction}

Scaling up data is paramount for robust and generalizable manipulation policies~\cite{o2024openx,luo2025beingh0, rise2, data_scaling_law, pi0}. 
However, the field of robotics suffers from a severe data scarcity bottleneck: acquiring robotic demonstrations is labor-intensive and costly~\cite{rh20t, robomind, droid}. 
In contrast, human demonstrations offer a readily accessible, virtually unlimited source of diverse real-world interaction data~\cite{liu2024taco,zhan2024oakink2,yang2022oakink,chao2021dexycb, ego4d}. 
Learning from these human videos presents a promising avenue towards general-purpose embodied AI.

To harness this vast potential of human data for robotic policy learning, prior work has explored a range of cross-embodiment transfer strategies, including visual editing~\cite{rise2, li2025h2r, lepert2025masquerade, phantom, ar2d2}, unified representation learning~\cite{ kareer2025pi05ego, luo2025beingh0, egomimic, humanpolicy, egovla, bu2025univla, lapa, moto, mimicplay}, and object-centric learning~\cite{chen2025vividex, vidbot, zeromimic, hrb}. However, these methods remain limited by the intrinsic discrepancy between human and robot demonstrations in visual appearance, 3D geometry, and embodiment action semantics. Consequently, they often introduce visual artifacts, learn representations with kinematic inconsistency to robot actions, or rely on brittle object state estimation pipelines. These limitations substantially reduce transfer efficiency and robustness in real-world manipulation.

To this end, we propose \model (shown in Fig. \ref{fig:teaser}), a human-to-robot imitation \textbf{L}earning paradigm that bridges the embodiment gap through \textbf{I}mplicit feature \textbf{D}istillation and \textbf{E}xplicit geometric \textbf{A}lignment. Our approach tackles the alignment problem from two complementary aspects:
\begin{enumerate}
    \item[\textbf{(1)}] \textbf{Implicit equivalence in the 2D visual domain}. We introduce a transitive distillation pipeline that enforces \textbf{feature-level equivalence} between human and robot observations of semantically equivalent interactions.
    \item[\textbf{(2)}] \textbf{Explicit correspondence in the 3D geometric domain}. We propose an embodiment-agnostic alignment strategy by \textbf{filtering} embodiment-specific geometries and \textbf{filling} an equivalent gripper.
\end{enumerate}

For \textbf{(1)}, realizing this visual alignment is inherently challenging due to the shortage of strictly-paired human-robot data, as it requires precise geometric correspondence between human and robot when manipulating the same objects.
Therefore, we construct an intermediate ``pseudo-robot'' domain as a transitive feature bridge between human and robot observations.
Specifically, we first construct pseudo-robot data $\mathcal{A}$ from human images by preserving equivalent interactions while replacing the human embodiment with a robot proxy. Next, leveraging the robot URDF model, we further synthesize pseudo-robot data $\mathcal{B}$ that shares an identical kinematic configuration with the real robot. 
Although $\mathcal{A}$ and $\mathcal{B}$ are not directly paired, they establish two equivalent correspondences: \emph{human-$\mathcal{A}$} and \emph{$\mathcal{B}$-real-robot}. 
Building upon this structure, we design a \textbf{transitive distillation} pipeline that propagates foundation visual representations~\cite{dinov3} from human to robot observations via the pseudo-robot domain.
After distillation, downstream visuomotor policies operate on feature embeddings that are numerically aligned across human and robot data. To power this transitive feature bridge, we construct the \textbf{HPP-5M} (\textbf{H}uman-to-\textbf{P}seudo-Robot \textbf{P}air) dataset, comprising nearly 5 million strictly paired, interaction-equivalent human and pseudo-robot frames, providing large-scale supervision for robust semantic alignment across diverse tasks and scenes.

For \textbf{(2)}, our 3D explicit alignment addresses geometric mismatches by constructing a canonical observation space via embodiment-aware \textbf{filter-and-fill} strategy. 
Specifically, we filter embodiment-specific geometries from both human and robot point clouds
and fill a virtual floating gripper as an embodiment-equivalent interaction proxy into the remaining scene.
This design explicitly decouples embodiment from interaction geometry, enabling consistent 3D-aware perception for cross-embodiment policy transfer.

We empirically validate \model in terms of both data efficiency and OOD robustness.
Importantly, our work focuses on improving human data utilization rather than introduce a new visuomotor architecture. To isolate this effect, we adopt a recent powerful diffusion-based 3D policy with DINOv3~\cite{dinov3} feature injection, RISE-2~\cite{rise2}, as the fixed baseline across all experiments.
To evaluate data efficiency, we compare policies trained on mixed human-robot datasets with those trained on varying amounts of robot-only data. 
Results show that human data substitutes up to 80\% of costly robot demonstrations, enabling effective policy learning with minimal robot supervision.
To assess the out-of-distribution (OOD) generalization, we study scenarios involving novel objects and visual distractors introduced through human data. 
\model successfully transfers the unseen patterns from human videos, whereas robot-only baseline exhibits significant performance degradation.

Our main contributions are summarized as follows:
\begin{itemize}
    \item We present \model, a paradigm that bridges the embodiment gap by jointly combining \emph{implicit feature distillation} and \emph{explicit geometric alignment}, enabling policy to benefit from human demonstrations. 
    \item We enable effective knowledge transfer from human to robot observations by establishing a shared feature space that preserves equivalent interactions across embodiments.
    \item We achieve embodiment-agnostic geometric perception by decoupling interaction structure from the acting agent, yielding consistent 3D-aware representations for reliable policy transfer.
    \item We introduce HPP-5M, a dataset comprising nearly \emph{5 million} interaction-equivalent human-to-pseudo-robot frame pairs from diverse human manipulation videos, which enables scalable semantic alignment between human and robot observations.

\end{itemize}

\begin{figure*}[!htp]
    \centering
    \includegraphics[width=\textwidth]{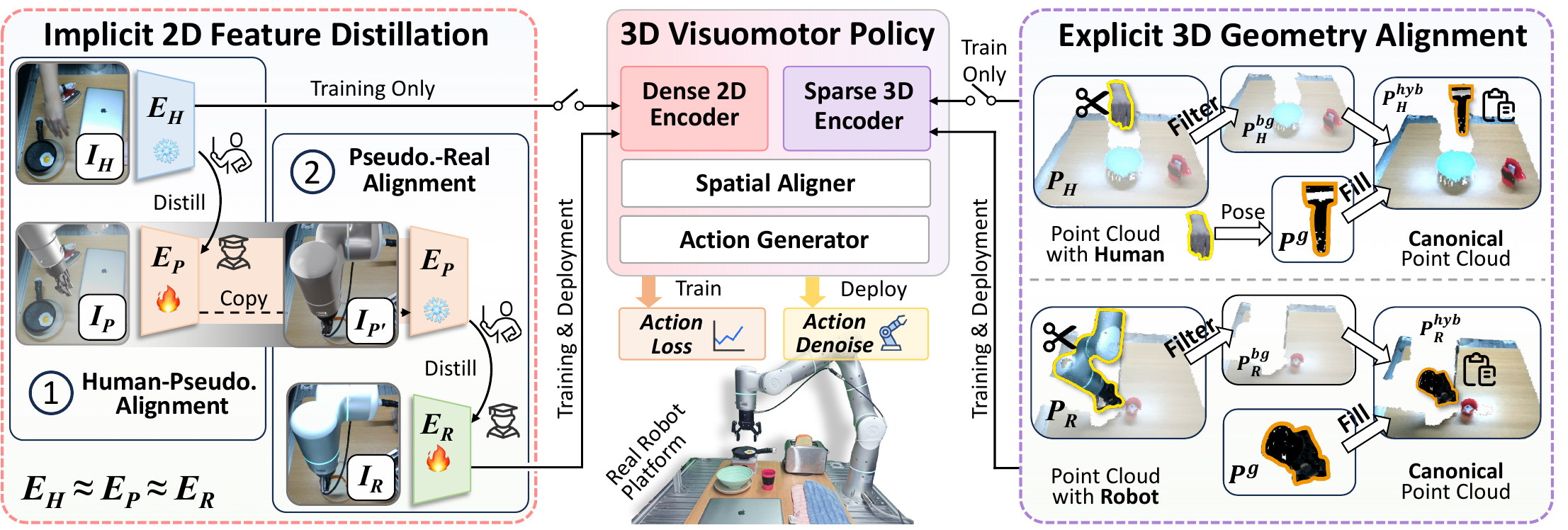}
    \caption{\textbf{The \model Framework.} (\textit{Left}) Stage \textcircled{1} establishes semantic equivalence by distilling features from human observations to pseudo-robot counterparts. Stage \textcircled{2} then trains the real-robot encoder to match the pseudo-robot representations, achieving a shared latent space where $E_H \approx E_P \approx E_R$. (\textit{Right}) To construct a canonical 3D observation space, embodiment-specific geometries are filtered from the unprojected point clouds. A virtual gripper is then filled into the scene, yielding hybrid 3D observations. (\textit{Center}) The 3D visuomotor policy fuses these aligned dense 2D features and sparse 3D tokens to predict continuous actions.}
    \vspace{-1em}
    \label{fig:method}
\end{figure*}

\section{RELATED WORKS}
\label{sec:related_works}

\subsection{Learning from Human Videos}

Leveraging human video data is promising for scalable robot learning, but progress is limited by the cross-embodiment gap. The direct approach utilizes visual editing~\cite{rise2, li2025h2r, lepert2025masquerade, phantom, ar2d2} to translate human videos into "robot videos" via pixel-level inpainting. Although effective, these methods are computationally expensive, often introduce visual artifacts, and typically operate in 2D, failing to preserve the consistent 3D geometry needed for depth-aware policies.

Another direction studies unified representation learning, training policies directly on human data via co-training~\cite{kareer2025pi05ego, luo2025beingh0, egomimic, humanpolicy, egovla} or unified tokenization~\cite{bu2025univla, lapa, moto, mimicplay}. However, substantial kinematic mismatch between humans and robots challenges these approaches~\cite{x-diffusion}. Auxiliary objectives, such as predicting human hand parameters~\cite{luo2025beingh0}, do not map cleanly to robot end-effectors, limiting transfer efficiency and requiring substantial fine-tuning. Meanwhile, object-centric methods~\cite{chen2025vividex, vidbot, zeromimic, hrb} bypass the embodiment gap by extracting affordances or retargeting object trajectories. Yet, they rely on explicit state estimation, which remains ambiguous for category-level objects and difficult for deformable or articulated objects, limiting applicability in complex environments.

In contrast, \model avoids the pitfalls of pixel-level editing, kinematic inconsistencies, and ambiguous state estimation. By combining implicit feature distillation with explicit geometric alignment to mitigate the embodiment gap, \model enables effective learning from human videos and efficient transfer to robots.

\subsection{Visual Representation Learning in Robotics}
Visual perception enables embodied agents to interpret and interact with unstructured environments. Early imitation learning~\cite{zhang2018deep,pari2021surprising} relied on pre-trained visual representations to extract essential features for robot control~\cite{r3m, liv, vip, mvp}. While effective in constrained settings, they lack sufficient generalization to diverse real-world environments, and their performance degrades substantially due to the domain gap between pre-training and deployment.

Visual foundation models~\cite{clip,dinov2,dinov3,sd}, trained via self-supervised learning (SSL) on internet-scale images, significantly improve representation robustness and transferability~\cite{cage}. Consequently, recent research adapts these models for robotics via two main directions: distilling broad representations into robotics-tailored features~\cite{theia, deng2026robot, gnfactor}, and incorporating them directly into visuomotor policies~\cite{rise2, cage, same, spawnnet, soft}.

Inspired by their strong transferability, \model\ distills DINOv3~\cite{dinov3} to learn a representation space aligning human and robot observations. Unlike standard SSL matching different views of the same image, we use semantically equivalent observations across embodiments as the distillation signal, reducing the visual domain gap.

\section{Method}
\label{sec:method}

The \model framework aims to enable data-efficient learning of 3D visuomotor policies from human video demonstrations. This approach reduces the reliance on extensive robotic teleoperation while avoiding the visual artifacts introduced by pixel-level editing. As illustrated in Fig.~\ref{fig:method}, the pipeline consists of three components: \emph{2D Feature Distillation}, \emph{3D Geometric Alignment}, and \emph{Mixed-Data Policy Learning}.

\subsection{Implicit 2D Feature Distillation} 
\label{sec:feat_distillation}

We formulate 2D feature alignment as a two-stage \emph{transitive distillation} problem where human observations are first aligned with a pseudo-robot intermediate domain and subsequently transferred to real-robot observations. 
Instead of directly matching human and robot observations, this intermediate bridge enables representation alignment in a shared latent space.

\subsubsection{\textbf{Stage 1: Human-to-Pseudo-Robot Distillation}}  
Directly aligning human observations with real-robot data is infeasible due to the scarcity of strictly paired demonstrations. 
However, by replacing the human embodiment with a robot proxy (pseudo-robot) through visual editing, we can construct a pseudo-robot domain in which the human and robot can have equivalent interaction. 
This stage therefore establishes semantic equivalence between human hands and robotic grippers within this domain.

We form paired observations $\{I_H, I_P\}$, where $I_H$ denotes human demonstration frames and $I_P$ denotes their pseudo-robot counterparts synthesized offline. 
Let $E_H$ and $E_P$ denote the two visual encoders for $I_H$ and $I_P$, respectively, both initialized from pre-trained DINOv3 weights~\cite{dinov3}.
During distillation, $E_H$ is frozen as the teacher while $E_P$ is trained to match teacher's representations. 
Following DINO-style self-distillation~\cite{dino,dinov2,dinov3}, training relies on multi-crop augmentation to enforce semantic-invariant representations.
However, under cross-embodiment settings, standard random local crops often emphasize background correlations rather than interaction. 
To preserve interaction equivalence across human and robot, we replace random local crops with a Region-of-Interaction (RoInt) cropping strategy that restricts local views to centered around hand/gripper-object contact regions.
During distillation, the teacher always receives full images, while the student is occasionally exposed to RoInt-cropped views. 
This process encourages the student encoder to capture the embodiment-agnostic interaction semantics from pseudo-robot images.

\subsubsection{\textbf{Stage 2: Robot Pseudo-to-Real Distillation}}
The second stage transfers the aligned representations from the pseudo-robot domain to real robot observations, completing the transitive bridge toward real-world deployment. 

Given paired observations $\{I_{P'}, I_R\}$, where $I_R$ denotes real-robot images and $I_{P'}$ their pseudo-robot counterparts (construction detailed below), we use $E_P$ from Stage 1 as the teacher encoder and $E_R$ as the student.
Both encoders are initialized with the optimized Stage 1 weights.
During distillation, $E_P$ remains frozen, while $E_R$ is trained to match the teacher representations.

Unlike Stage 1, which resolves embodiment discrepancies between human and robot, the agents in $I_{P'}$ and $I_R$ share identical physical topology, leaving only photometric differences from sim-to-real variations.
We therefore discard RoInt cropping and adopt standard DINO-style global and local multi-crop augmentation, enabling $E_R$ to align photometric statistics using full-image context.

Through this two-stage process, representations become transitively aligned across domains, $E_H \approx E_P \approx E_R$, establishing semantic equivalence from human demonstrations to real-robot observations.

\begin{figure}[!tbp]
    \centering
    \includegraphics[width=\linewidth]{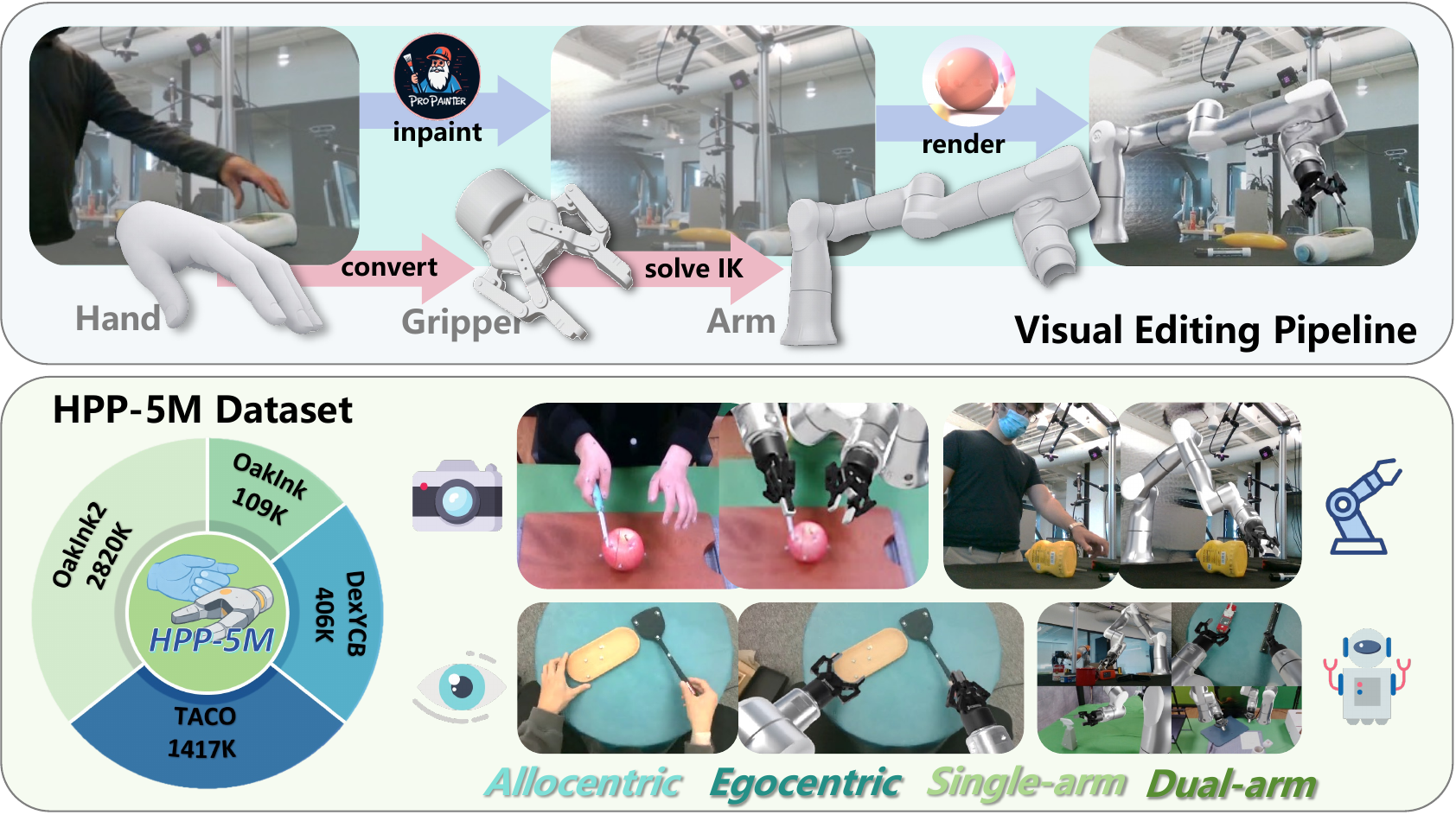}
    \caption{\textbf{Overview of the HPP-5M Dataset Generation and Composition.} \textit{(Top)} visual editing pipeline; \textit{(Bottom)} dataset illustration. }
    \label{fig:HPP}
    \vspace{-1em}
\end{figure}

\subsubsection{\textbf{Paired Data for Distillation}}
Both distillation stages require paired cross-embodiment observations, but the domain gap each stage should bridge -- and thus the data scale -- differs considerably.

\emph{i) Human-to-Pseudo Paired Data: HPP-5M.}
Because human hands and robotic grippers differ considerably in morphology and appearance, Stage 1 requires relatively large-scale paired data to bridge this visual feature gap.
We leverage existing 3D hand-object interaction datasets as the human-side source and construct the HPP-5M (Human-to-Pseudo-Robot Pair) dataset (Fig.~\ref{fig:HPP}) from DexYCB~\cite{chao2021dexycb}, TACO~\cite{liu2024taco}, OakInk~\cite{yang2022oakink}, and OakInk2~\cite{zhan2024oakink2}, all annotated with parametric 3D hand poses.
Human hand motions are converted into equivalent gripper actions by fitting the gripper wrist pose and opening state from fingertip 3D positions, solving arm configurations via inverse kinematics (IK), removing the human from the original image through generative inpainting~\cite{zhou2023propainter}, and finally rendering the pseudo-robot into the scene using a path-traced renderer.
This process produces strictly paired observations $\{I_H, I_P\}$ that preserve interaction equivalence across embodiments.
In total, HPP-5M comprises approximately 5M paired frames from 18K video sequences, 
spanning single-hand and bimanual interactions, single-object and object-object manipulations, and diverse allocentric and egocentric viewpoints.
Additionally, to provide scene-specific priors for real-world deployment, we incorporate a small amount of target-domain free-motion human data (23k frames, 0.4\% of HPP-5M). 
These task-agnostic human-play videos, collected on our robot platform, are converted into pseudo-robot observations via visual editing and appended to the Stage 1 training set.

\emph{ii) Pseudo-to-Real Paired Data.}
Stage 2 only needs to bridge the photometric gap between rendered and real images of the same robot, so a small set of free-move robot recordings suffices.
We record real-world robot joint trajectories and synthesize a pseudo-robot mesh under identical kinematic configurations.
The pseudo-robot mesh is rendered onto the native real-robot image $I_R$ to produce its pseudo-robot counterpart $I_{P'}$, forming geometry-identical pairs $\{I_{P'}, I_R\}$.

\subsubsection{\textbf{Distillation Objectives}} 
To optimize the student networks ($E_{P}$ in the stage 1 and $E_{R}$ in the stage 2), we formulate a combined self-supervised learning objective based on the DINOv3~\cite{dinov3}. Formally, our total distillation objective is defined as:
\begin{equation}
    \setlength{\abovedisplayskip}{3pt}
    \setlength{\belowdisplayskip}{3pt}
    \mathcal{L}_{\mathrm{total}} = \mathcal{L}_{\mathrm{DINO}} + \mathcal{L}_{\mathrm{iBOT}} + \lambda \mathcal{L}_{\mathrm{KoLeo}},
    \label{eq:distill_objective}
\end{equation}
where $\mathcal{L}_{\mathrm{DINO}}$ denotes the self-distillation loss applied to global CLS tokens,  enforce teacher and student's semantic consistency. $\mathcal{L}_{\mathrm{iBOT}}$ is a masked patch prediction objective that trains the student to reconstruct teacher patch features, providing dense supervision at local level, and $\mathcal{L}_{\mathrm{KoLeo}}$ is a differential entropy regularizer that promotes a uniform feature distribution across the batch, weighted by $\lambda$ (set to 0.1).
We deliberately omit the Gram loss, originally introduced to stabilize dense representations, as such degradation does not arise in our short-horizon cross-domain distillation.
When the student operates on local crops, only $\mathcal{L}_{\mathrm{DINO}}$ is applied, while $\mathcal{L}_{\mathrm{iBOT}}$ and $\mathcal{L}_{\mathrm{KoLeo}}$ are disabled.

\subsection{Explicit 3D Geometric Alignment}
\label{sec:geo_alignment}
Although implicit 2D distillation aligns human and robot observations at the feature level, this alignment alone is insufficient when human data is directly used to train depth-aware 3D policies. 
Such policies rely not only on semantic consistency but also on geometrically coherent observations for spatial modeling. 
Therefore, beyond feature-level alignment, we further establish cross-embodiment consistency at the geometric level by explicitly align the 3D observation space.

\subsubsection{\textbf{Embodiment-Specified Filtering}} 
Given a scene point cloud extracted from an RGB-D observation, we first remove embodiment-specific geometry to obtain an agent-independent scene representation.

For human demonstrations, where explicit kinematic descriptions (URDF) are unavailable, we apply visual segmentation. 
A foundation segmentation model, Grounded-SAM2~\cite{ren2024groundedsam}, extracts hand-arm masks from the RGB image, which are applied during depth unprojection to remove all human-associated points, producing a clean, human-free background point cloud $P_{H}^{bg}$.

For robot teleoperation data, we instead leverage proprioceptive measurements. 
Given real-time joint angles and the robot URDF model, forward kinematics determines the spatial configuration of all robot links. 
We then compute the robot occupancy volume by injecting the link geometries in 3D space. 
Points in the observed point cloud that fall within a predefined margin of this occupancy volume are identified as robot-specific observations and removed, yielding an arm-free scene point cloud $P_{R}^{bg}$.

After embodiment decoupling, both $P_{H}^{bg}$ and $P_{R}^{bg}$ reside in a shared 3D observation space containing only embodiment-agnostic background objects.

\subsubsection{\textbf{Canonical Gripper Filling}} 
While embodiment filtering removes discrepancies, it also leave the depth-aware policy ``blind" to the location and state of the end-effector. 
To restore interaction grounding without reintroducing embodiment-specific artifacts, we fill a canonical virtual floating gripper into the decoupled scenes $P_{H}^{bg}$ and $P_{R}^{bg}$.

We define a generic gripper point-cloud template parameterized by its opening state $s$, denoted as $P^{g}(s)$. 
For robot data, the Tool Center Point (TCP) pose $\mathbf{T}_R$ and gripper opening state $s_R$ are obtained directly from proprioception. 
For human demonstrations, we leverage a generalizable multi-view hand pose estimator, POEM~\cite{yang2025poemv2}, for 3D hand joints.
Fingertips are aligned to the virtual gripper tips via least-squares to optimize an equivalent TCP pose $\mathbf{T}_H$ and opening state $s_H$.

After spatial transformation into the camera frame, the canonical gripper point cloud is used to fill the decoupled background scenes, yielding the final hybrid 3D observations for both domains:
\begin{equation}
\small
    P_{H}^{hyb} = P_{H}^{bg} \cup \big(\mathbf{T}_{H} \cdot P^{g}(s_{H})\big), \;
    P_{R}^{hyb} = P_{R}^{bg} \cup \big(\mathbf{T}_{R} \cdot P^{g}(s_{R})\big).
\end{equation}
Through this explicit injection, the 3D policy observes an identical geometric ($P_{H}^{hyb}$ and $P_{R}^{hyb}$) across both human and robot data -- comprising purely the task-relevant environment and a unified end-effector.

\subsection{Policy Learning and Deployment}
With the visual semantics and 3D geometries strictly aligned across human and robot observation, we integrate these representations into a state-of-the-art 3D diffusion-based policy model, RISE-2~\cite{rise2}. 

\qheading{Training.} During the training phase, we adopt a mixed-data training paradigm utilizing both human demonstrations and a portion of robot demonstrations. For a given human data sample, the raw native RGB image $I_{H}$ is processed by a frozen, pre-trained human encoder $E_{H}$ (identical to DINOv3) to extract the dense 2D feature. Concurrently, the hybrid 3D observation $P_H^{hyb}$ is processed by the sparse 3D encoder of RISE-2 to extract sparse 3D geometric tokens.
In contrast, for a robot data sample, the raw robot image $I_{R}$ is fed into the frozen real-robot encoder $E_{R}$ (obtained from $E_{H}$ via transitive distillation) to obtain the corresponding dense 2D features, while $P_R^{hyb}$ is encoded by the shared Sparse 3D Encoder. 
$E_{H}$ and $E_{R}$ are inherently aligned in the latent space through our transitive distillation, and $P_H^{hyb}$ and $P_R^{hyb}$ share canonical geometry representation. The extracted 2D dense features and 3D sparse tokens are then fused and passed through the remaining transformer layers of the RISE-2 policy model to iteratively denoise action trajectories.

\qheading{Deployment.} 
During real-world deployment, the policy operates exclusively on the robot's real-time sensor streams without requiring generative visual editing. 
Given an RGB-D observation, the raw image $I_R$ is processed by $E_R$ to extract dense 2D features. Concurrently, the robot's spatial occupancy is directly filtered from the unprojected 3D point cloud to obtain the background $P_R^{bg}$. By injecting the canonical virtual gripper using the real-time TCP pose and gripper state, we form the hybrid 3D observation $P_R^{hyb}$ for the Sparse 3D Encoder. Because both 2D and 3D observation spaces are rigorously aligned with the training distribution, the policy directly fuses these features to predict continuous actions for receding horizon control.

\section{EXPERIMENTS}
\label{sec:exp}

\begin{figure*}[!t]
    \centering
    \includegraphics[width=\textwidth]{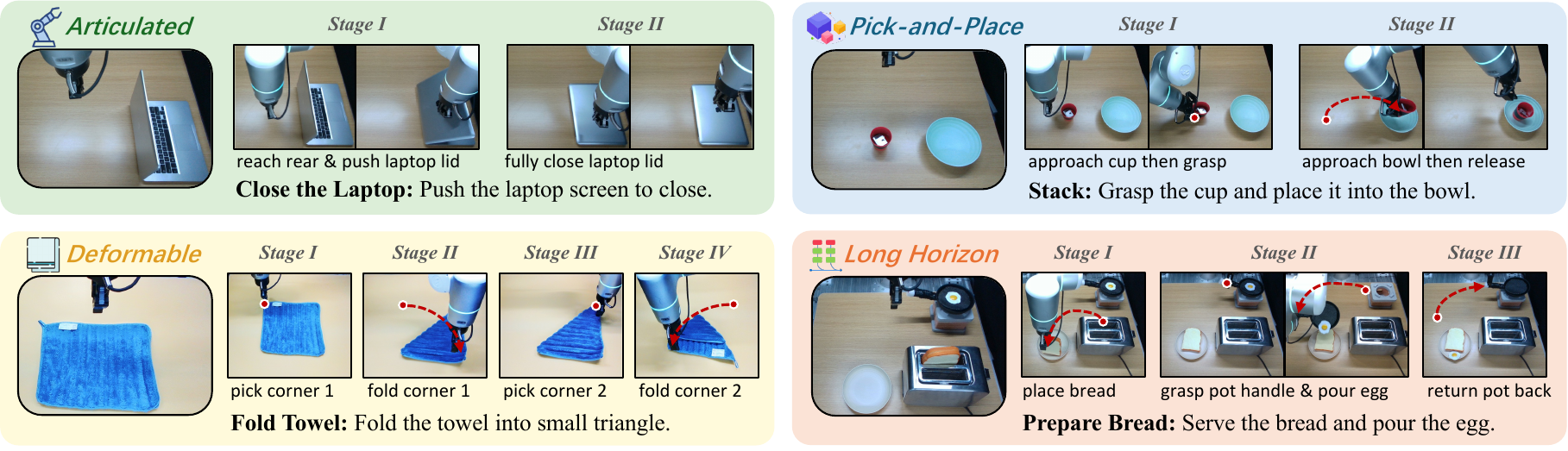}
    \caption{\textbf{Overview of 4 Real-World Manipulation Tasks.} We evaluate policies across four types of interactions: \textbf{\textit{Close Laptop}} (articulated), \textbf{\textit{Stack}} (6 DoF pick-and-place), \textbf{\textit{Fold Towel}} (deformable), and \textbf{\textit{Prepare Bread}} (long-horizon).}
    \label{fig:task}
\end{figure*}
\begin{figure*}[!t]
    \centering
    \includegraphics[width=\textwidth]{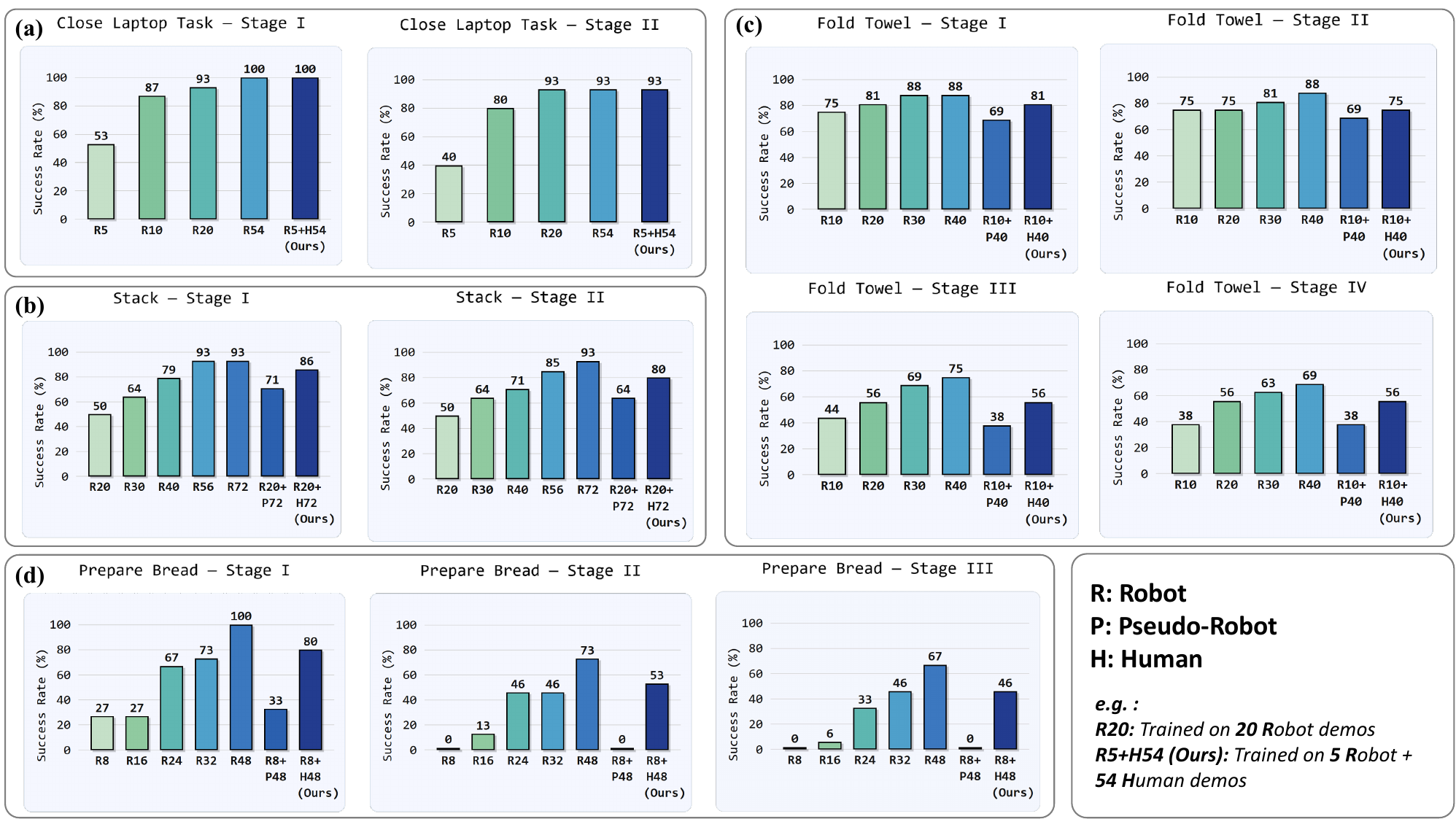}
    \caption{\textbf{Evaluation of Data Efficiency across 4 Real-World Manipulation Tasks.} The charts compare the success rates of policies trained on varying amounts of robot-only data, pseudo-robot baseline data, and mixed human-robot data.}
    \label{fig:main_results}
\end{figure*} 

\subsection{Setup}
\qheading{Hardware Platform.} 
Our real-world setup consists of a Flexiv Rizon 4 robotic arm equipped with a Robotiq 2F-85 gripper and a global Intel RealSense D415 RGB-D camera for scene perception. 
Real-world robot demonstrations are collected via haptic teleoperation. 

\qheading{Tasks.} We design four real-world manipulation tasks for evaluation: \textit{\textbf{Close Laptop}} (articulated-object), \textit{\textbf{Stack}} (6 DoF pick-and-place), \textit{\textbf{Fold Towel}} (deformable-object), and \textit{\textbf{Prepare Bread}} (long-horizon task).
Please refer to Fig.~\ref{fig:task} for task illustrations and progressive evaluation stages.

\qheading{Baseline.} To highlight the advantage of our implicit and explicit alignment approach, we establish a Pseudo-Robot baseline for comparison. It represents mainstream explicit visual editing methods~\cite{phantom, lepert2025masquerade, li2025h2r}, where human images are edited into pseudo-robot images to train a same 3D visuomotor policy. 

\subsection{Data Efficiency}

We evaluate the data efficiency of \model across four manipulation tasks to \textbf{quantify the equivalence of human data to robot data}. 
The results are plotted in Fig.~\ref{fig:main_results}.

In tasks requiring simple motion primitives and coarse geometric constraints (e.g., \textbf{\textit{Close Laptop}}), our approach efficiently extracts motion patterns (\textit{e.g.}, pushing the laptop lid along its hinge axis) directly from human demonstrations. 
In 6 DoF pick-and-place task like \textbf{\textit{Stack}}, the model effectively transfers spatial coordination and grasp timing learned from human data, enabling accurate object alignment and placement.

While these results demonstrate the data efficiency enabled by human demonstrations, they do not explain \emph{why} such transfer succeeds. 
We therefore compare our method with explicit visual-editing baselines \cite{phantom,lepert2025masquerade}, which directly translates human observations into pseudo-robot images, to isolate the effect of geometric alignment.
Although such methods capture coarse motion patterns, they exhibit noticeable performance degradation in tasks requiring precise 3D localization. 
For instance, manipulating deformable objects, \textbf{\textit{Fold Towel}}, is highly sensitive to precise spatial localization, where noisy depth introduced by pseudo-robot rendering corrupts the fine-grained spatial perception required for reliable grasping. 
In contrast, \model canonicalizes the 3D observation through geometric alignment, providing consistent human–robot perception when trained on mixed data.

We further examine capability transfer in the challenging \textbf{\textit{Prepare Bread}} task, a long-horizon manipulation requiring sequential execution under diverse spatial configurations, where the positions of the toaster, plate, and stove vary across trials. 
A policy trained on only 8 robot demonstrations struggles to generalize. 
However, augmenting these demonstrations with human data (\textit{8 Robot + 48 Human}) enables the transfer of stable grasping and 6-DoF manipulation, improving success rates to 80\%, 53\%, and 46\% across the three stages.
However, this gain does not emerge when human data are incorporated through explicit pixel-level visual editing (33\%, 0\%, and 0\% across three stages). 
The pseudo-robot baselines (\textit{e.g.}~\cite{phantom,lepert2025masquerade}) introduces artifacts and depth inconsistencies that accumulate over the long horizon, disrupting the stable 3D perception required for sequential manipulation.
In contrast, \model combines implicit feature distillation with explicit 3D geometric alignment, suppressing these geometric confounders and making human-to-robot transfer reliable for multi-stage execution.

\subsection{Out-of-Distribution Generalization}

We evaluate whether \model transfers OOD robustness from human videos to robotic execution on \textbf{\textit{Fold Towel}}.
We train with strictly in-domain robot demonstrations that use a standard blue towel, but test under an OOD configuration: the target is a novel pink towel, and a folded blue towel is placed nearby as a strong visual distractor.

\begin{table}[htbp]
\centering
\begin{tabular}{lcccc}
\toprule
\textbf{Method (OOD Setting)} & \textbf{I} & \textbf{II} & \textbf{III} & \textbf{IV} \\
\midrule
40 Robot & 36 & 27 & 18 & 18 \\
\textbf{40 Robot + 40 Human (Ours)} & \textbf{63} & \textbf{54} & \textbf{27} & \textbf{27} \\
\bottomrule
\end{tabular}
\vspace{1ex}
\caption{Out-of-Distribution Generalization on the \textbf{\textit{Fold Towel}} task. The evaluation introduces a novel pink towel and a folded blue towel as a distractor. Stage I to IV represent the four steps of the folding task. Success rates (\%) are reported.}
\label{tab:ood_results}
\end{table}

As shown in Table~\ref{tab:ood_results}, policies trained only on in-domain robot data (\textit{40 Robot}) fail to generalize, reaching only 18\% final success in the OOD setting. 
The distractor particularly harms the initial corner-grasping stage: the in-domain blue towel draws attention away from the functional target, reducing first-corner success to 36\% and cascading into downstream failures.

In contrast, augmenting training with OOD human demonstrations (\textit{40 Robot + 40 Human}) substantially improves robustness, increasing first-corner success to 63\% and final completion to 27\%. 
These gains indicate that \model leverages diverse human videos to learn appearance-robust interaction cues, allowing the policy to identify the functional target and ignore strong distractors without requiring extensive OOD robot data collection.

\subsection{Ablation}

To evaluate the critical components of our \model framework, we conduct ablation studies on the \textbf{\textit{Stack}} task using the \textit{20 Robot+72 Human} data mixture. We systematically analyze our design choices within the implicit 2D distillation and explicit 3D alignment pipeline.

\begin{table}[htbp]
\centering
\begin{tabular}{lcc}
\toprule
\textbf{Method (2D Alignment)} & \textbf{I (Grasp)} & \textbf{II (Place)} \\
\midrule
\textbf{20 Robot + 72 Human (Ours)} & \textbf{86} & \textbf{80} \\
w/ DINOv3 (No Distillation) & 20 & 20 \\
w/ Stage-1 Distillation Only & 67 & 60 \\
w/o Internet Pre-training & 73 & 67 \\
w/o Free-motion Pre-training & 33 & 33 \\
\bottomrule
\end{tabular}
\vspace{1ex}
\caption{Ablation study of implicit 2D feature distillation on the \textbf{\textit{Stack}} task. 
We analyze the contribution of each component in the dual-stage alignment pipeline and the impact of different pre-training data sources on policy performance.}\vspace{-1em}
\label{tab:ablation_2d}
\end{table}
\qheading{Implicit 2D Distillation.} Table~\ref{tab:ablation_2d} analyzes the necessity of our dual-stage feature distillation and dataset composition design. 
Firstly, applying a pre-trained DINOv3 encoder without cross-embodiment distillation (\textit{w/ DINOv3}) leads to near-total failure (20\%). This proves that general visual representations encode human and robot arm as distinct semantic entities, preventing effective transfer without latent alignment.

Additionally, relying solely on the first distillation stage (\textit{w/ Stage-1 Distillation Only}) causes a noticeable performance drop (19\% and 20\% decrease). Stage 1 aligns human images with pseudo-robot images but leaves a critical Pseudo-to-Real gap, which our Stage 2 distillation successfully bridges. 

Pre-training data composition is also important. Removing large-scale internet data (\textit{w/o Internet Pre-training}) restricts representation diversity, dropping performance by 13\%. Removing target-domain free-motion data (\textit{w/o Free-motion Pre-training}) collapses performance to 33\% (53\% and 47\% decrease), as internet-only data lacks the scene-specific priors necessary for actual physical deployment.

\begin{table}[htbp]
\centering
\begin{tabular}{lcc}
\toprule
\textbf{Method (3D Alignment)} & \textbf{I (Grasp)} & \textbf{II (Place)} \\
\midrule
\textbf{20 Robot + 72 Human (Ours)} & \textbf{86} & \textbf{80} \\
w/o Filter Human & 66 & 60 \\
w/o Filter Robot & 53 & 53 \\
w/o Filter Both & 40 & 40 \\
\bottomrule
\end{tabular}
\vspace{1ex}
\caption{Ablation on explicit 3D geometric alignment in the \textit{\textbf{Stack}} task. ``Filter" refers to the process of decoupling the specific embodiment (removing the human hand or robot arm point cloud) and filling the unified virtual gripper.}
\vspace{-1em}
\label{tab:ablation_3d}
\end{table}
\textbf{Explicit 3D Alignment.} 
Table~\ref{tab:ablation_3d} demonstrates the indispensable role of explicitly decoupling the agent's geometry in the 3D observation space. When neither point clouds are filtered (\textit{w/o Filter Both}), success rates drop by 46\% and 40\%. This performance is worse than the pure robot baseline (10\% decrease), indicating mixing 3D geometries without alignment causes severe negative transfer.

Furthermore, asymmetric filtering introduces fatal training-deployment mismatches. Retaining only the robot point cloud (\textit{w/o Filter Robot}) confronts the policy with massive, unedited robot arm point clouds during deployment, dropping success rates by 33\% and 27\%. Retaining only the human hand point cloud (\textit{w/o Filter Human}) introduces out-of-distribution geometric noise, causing 20\% decrease. These results confirm that constructing a strictly embodiment-agnostic 3D space by filtering native geometries and filling a unified virtual gripper is essential for cross-embodiment learning.

\begin{figure}[!tbp]
    \centering
    \includegraphics[width=\linewidth]{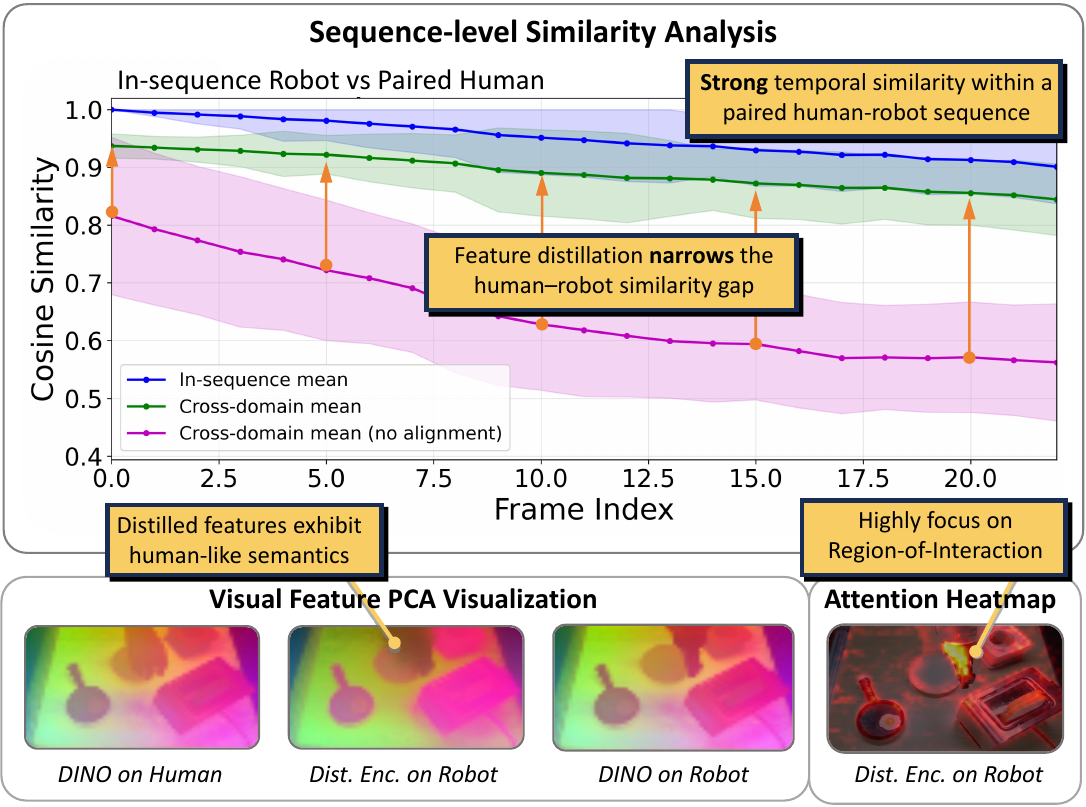}
    \caption{\textbf{Empirical Analysis of the Feature Distillation.} (\textit{Top}) sequence-level similarity; (\textit{Bottom}) PCA of the aligned feature.}\vspace{-1em}
    \label{fig:feature_alignment}
\end{figure}
\subsection{Analysis of 2D Feature Distillation}

To validate that our implicit 2D feature distillation effectively bridges the cross-embodiment gap, we analyze the learned latent space both quantitatively and qualitatively. 

For sequence-level similarity analysis, we evaluate whether aligned human features preserve the temporal structure of robot demonstrations. 
Specifically, we compute cosine similarity between the global feature of the \textit{first frame} and \textit{subsequent frames} within each paired sequence.
We conduct three comparisons to analyze different alignment conditions.
The blue curve (\emph{In-sequence mean}) serves as an upper bound, measuring similarity between the first pseudo-robot frame (reference) and later pseudo-robot frames from the same sequence ($E_P(I_{P_1}) \leftrightarrow E_P(I_{P_{1:t}})$), thereby capturing intrinsic frame-level similarity and temporal consistency in robot observations.
The purple curve (\emph{Cross-domain mean (no alignment)}) replaces the reference with a paired first human frame extracted using original DINO encoders and compares it against pseudo-robot features extracted by the same DINO encoder (unaligned), revealing the substantial human–robot feature gap ($\mathrm{DINO}(I_{H_1}) \not\sim \mathrm{DINO}(I_{P_{1:t}})$).
The green curve (\emph{Cross-domain mean}) applies our aligned encoders under the same protocol ($E_H(I_{H_1}) \approx E_P(I_{P_{1:t}})$). 
The resulting similarity closely follows the in-sequence trend, demonstrating that feature distillation substantially narrows the human–robot gap while preserving temporal consistency across embodiments (Fig.~\ref{fig:feature_alignment} top).

Furthermore, Principal Component Analysis (PCA) of human and robot features (Fig.~\ref{fig:feature_alignment} bottom) reveals that the aligned encoder learns geometry-consistent representations, assimilating the robot end-effector into a human-like semantic space. 
Additionally, the self-attention heatmap shows our aligned robot encoder focuses on the Region-of-Interaction, indicating a stronger concentration of attention on the task-relevant execution region.

\section{Discussion}
\label{sec:discussion}

In this work, \model demonstrates that complex manipulation capabilities can be effectively transferred by combining a small amount of robot demonstrations with diverse human videos. However, a limitation lies in our explicit 3D alignment, which maps the highly articulated human hand to an equivalent parallel gripper. While effective for standard tasks, human hand geometry naturally aligns better with multi-fingered dexterous robotic hands. 
Future work will extend this framework in two directions:
1) supporting dexterous manipulation with multi-fingered hands, and
2) using the aligned visual encoder to train VLA or video-action models from human demonstrations.

{\balance
\printbibliography
}

\end{document}